\documentclass[accepted]{uai2022}  

\usepackage[american]{babel}

\usepackage{natbib} 
\usepackage{mathtools} 
\usepackage{booktabs} 
\usepackage{hyperref}       
\usepackage{url}            
\usepackage{nicefrac}       
\usepackage{amsfonts}		
\usepackage{graphicx}		
\usepackage{caption}		
\usepackage{longtable}		
\usepackage[ruled, linesnumbered, commentsnumbered]{algorithm2e}

\title{Test for non-negligible adverse shifts}

\author{
	Vathy M. Kamulete
	\href{https://orcid.org/0000-0002-4451-3743}{\includegraphics[scale=0.06]{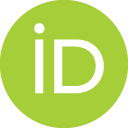}}
	\\
	Enterprise Model Risk Management \\
	Royal Bank of Canada \\
	Toronto, Canada \\
	\texttt{\href{mailto:vathy.kamulete@rbccm.com}{\nolinkurl{vathy.kamulete@rbccm.com}}} \\
}

\begin{document}
\maketitle

\begin{abstract}
  Statistical tests for dataset shift are susceptible to false alarms:
  they are sensitive to minor differences when there is in fact adequate
  sample coverage and predictive performance. We propose instead a
  framework to detect adverse shifts based on outlier scores,
  \texttt{D-SOS} for short. \texttt{D-SOS} holds that the new (test)
  sample is not substantively worse than the reference (training) sample,
  and not that the two are equal. The key idea is to reduce observations
  to outlier scores and compare contamination rates at varying weighted
  thresholds. Users can define what \emph{worse} means in terms of
  relevant notions of outlyingness, including proxies for predictive
  performance. Compared to tests of equal distribution, our approach is
  uniquely tailored to serve as a robust metric for model monitoring and
  data validation. We show how versatile and practical \texttt{D-SOS} is
  on a wide range of real and simulated data.
\end{abstract}

\section{Introduction}\label{sec:intro}
Suppose we fit a predictive model on a training set and predict on a test set.
Dataset shift, also known as data or population drift, occurs when training and
test distributions are not alike \citep{kelly1999impact, quionero2009dataset}.
This is essentially a sample mismatch problem. Some
regions of the data space are either too sparse or absent during
training and gain importance at test time. We want methods that alert
users to the presence of unexpected inputs in the test set
\citep{rabanser2019failing}. To do so, a measure
of divergence between training and test set is required. Can we not
use the many modern off-the-shelf multivariate tests of equal
distributions for this?

One reason for moving beyond tests of equal distributions is that they
are often too strict. They require high fidelity between training and
test set everywhere in the input domain. However, not \emph{all} changes
in distribution are a cause for concern -- some changes are benign.
Practitioners distrust these tests because of false alarms.
\citet{polyzotis2019data} comment:

\begin{quote}
	statistical tests for detecting changes in the data distribution
	{[}\ldots{]} are too sensitive and also uninformative for the typical
	scale of data in machine learning pipelines, which led us to seek
	alternative methods to quantify changes between data distributions.
\end{quote}

Even when the difference is small or negligible, tests of equal
distributions reject the null hypothesis of no difference. Monitoring
model performance and data quality is a critical part of deploying safe
and mature models in production \citep{paleyes2020challenges,
zhang2020machine}. An alarm should only be raised if a shift warrants
intervention. Retraining models when distribution changes are benign is 
both costly and ineffective \citep{vovk2021retrain}. To
tackle these challenges, we propose \texttt{D-SOS} instead.

In comparing the test set to the training set, \texttt{D-SOS} pays more
attention to the regions --- typically, the outlying regions --- where
we are most vulnerable. To confront false alarms, it uses a robust test
statistic, the weighted area under the receiver operating characteristic
curve (WAUC).
The weights in the WAUC discount the safe regions of the distribution.
As best we know, this is the first time that the WAUC is being used as a
test statistic in this context. The goal of \texttt{D-SOS} is to detect
\emph{non-negligible adverse shifts}. This is reminiscent of
noninferiority tests \citep{wellek2010testing}, widely used in
healthcare to determine if a new treatment is in fact not inferior to an
older one. Colloquially, \texttt{D-SOS} holds that the new sample is not
substantively worse than the old sample, and not that the two are equal.

\texttt{D-SOS} moves beyond tests of equal distributions and lets users
specify which notions of outlyingness to probe. The choice of the score
function plays a central role in formalizing what we mean by
\emph{worse}. These scores can come from out-of-distribution (outlier)
detection, two-sample classification, uncertainty quantification,
residual diagnostics, density estimation, dimension reduction, and more.
While some of these scores are underused and underappreciated in
two-sample statistical tests, they can be more telling than the
density-based scores. 

In this paper, we make the following contribution. We derive \texttt{D-SOS},
a novel and robust two-sample test for no adverse shift, from tests of
goodness-of-fit. The main takeaway is that given a generic method to assign
an outlier score to a data point, \texttt{D-SOS} turns these scores into a
two-sample test for no adverse shift. It converts arbitrary outlier
scores into interpretable probabilities, namely \(p-\)values. The field is
replete with tests of equal distribution and goodness-of-fit. We have
comparatively fewer options for tests of no adverse shift. We have created an 
accompanying \href{https://cran.r-project.org/package=dsos}{R package}
\texttt{dsos} for our method. In addition, all code and data used in this 
paper are publicly available.

\section{Motivational example}\label{sec:iris}

For illustration, we apply \texttt{D-SOS} to the canonical \texttt{iris}
dataset \citep{anderson1935irises}. The
task is to classify the species of Iris flowers based on \(d=4\)
covariates (features) and \(n=50\) observations for each species. 
We show how \texttt{D-SOS} helps diagnose false
alarms. We highlight that (1) changes in distribution do not necessarily
hurt predictive performance, and (2) points in the densest regions of
the distribution can be the most difficult -- unsafe -- to predict.

We consider four tests of no adverse shift. Each test uses a different
score. For two-sample classification, this score is the probability of
belonging to the test set. For density-based out-of-distribution (OOD)
detection, the score comes from isolation forest; this score is
inversely related to the local density. For residual diagnostics, it is
the out-of-sample (out-of-bag) prediction error from random forests.
Finally, for confidence-based OOD detection, it is the standard error of the
mean prediction from random forests, a proxy for prediction (resampling) 
uncertainty similar to \citep{schulam2019can}. Only
the first notion of outlyingness -- two-sample classification --
pertains to modern tests of equal distributions; the others capture
other meaningful notions of adverse shifts. For all these scores, higher
is \emph{worse}: higher scores indicate that the observation is
diverging from the desired outcome or that it does not conform to the
training set.

For the subsequent tests, we split \texttt{iris} into 2/3 training and
1/3 test set. The train-test pairs correspond to two partitioning
strategies: (1) random sampling and (2) in-distribution (most dense) examples
in the test set. How do these sample splits fare with respect to the
aforementioned tests? Let \(s\) and \(p\) denote \(s-\)value and \(p-\)value. The
results are reported on the \(s = -\, \log_{2}(p)\) scale because it is
intuitive and lends itself to comparison. We return to the advantages of
using \(s-\)value for comparison later. An \(s-\)value of \(k\) can be
interpreted as seeing \(k\) independent coin flips with the same outcome
-- all heads or all tails -- if the null is that of a \emph{fair} coin
\citep{greenland2019valid}. This conveys how incompatible the data is with
the null.

\begin{figure*}
	\centering
	\includegraphics[width=\textwidth]{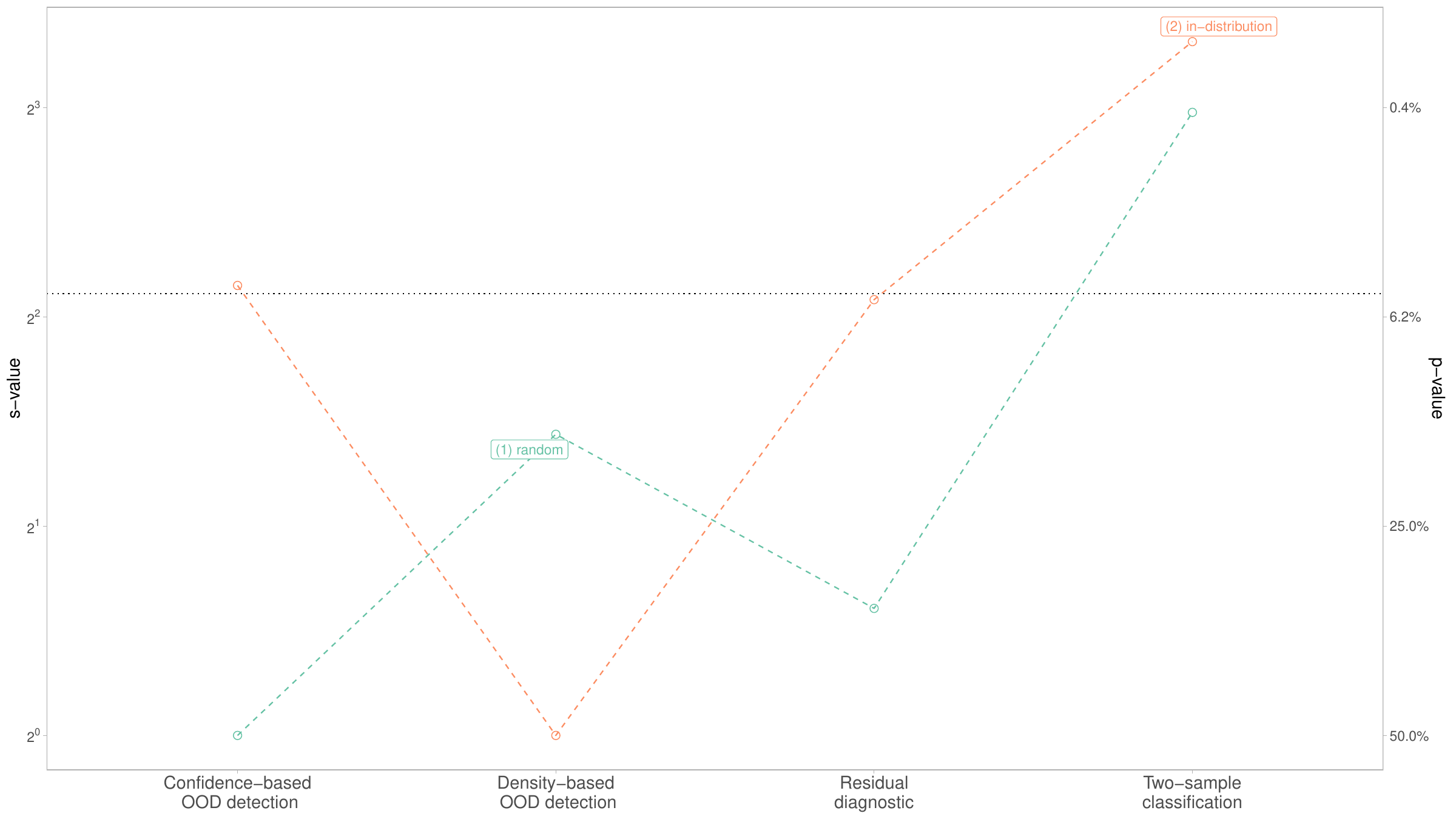}
	\caption{Tests of no adverse shift for \texttt{iris}. The tests cover 4
		notions of outlyingness for two train-test splits. The x-axis indicates
		the test type and the colour, the sampling (splitting) strategy. The
		dotted black line is the common, if not commonly abused, $p-$value $= 0.05$
		threshold. We clip \(s-\)values to a low and high of 1 and 10
		respectively and display a secondary y-axis with the \(p-\)value as
		a cognitive bridge.}
	\label{fig:iris-dsos}
\end{figure*}

In Figure \ref{fig:iris-dsos}, the case with (1) random sampling
exemplifies the type of false alarms we want to avoid. Two-sample
classification, standing in for tests of equal distributions, is
incompatible with the null of no adverse shift (a \(s-\)value of around
8). But this shift does not carry over to the other tests. Residual
diagnostics, density-based and confidence-based OOD detection are all
fairly compatible with the view that the test set is not worse. Had we
been entirely reliant on two-sample classification, we may not have
realized that this shift is essentially benign. Tests of equal
distributions alone give a narrow perspective on dataset shift.

Returning to Figure \ref{fig:iris-dsos}, density-based OOD detection
does not flag the (2) in-distribution test set as expected. We might be
tempted to conclude that the in-distribution observations are
\emph{safe}, and yet, the tests based on residual diagnostics and
confidence-based OOD detection are fairly incompatible with this view.
Some of the densest points are concentrated in a region where the
classifier does not discriminate well: the species `versicolor' and
`virginica' overlap. That is, the densest observations are not
necessarily safe. Density-based OOD detection glosses over this: the
trouble may well come from inliers that are difficult to predict. We get
a more holistic perspective of dataset shift because of these
complementary notions of outlyingness.

\section{Statistical framework}\label{sec:test-statistic}

The theoretical framework builds on \citet{zhang2002powerful}. Take an i.i.d.
training set \(X^{tr} = \{x_{i}^{tr}\}_{i=1}^{n^{tr}}\) and a test set
\(X^{te} = \{x_{i}^{te}\}_{i=1}^{n^{te}}\). Each dataset \(X^{o}\) with
origin \(o \in \{tr(ain), te(st)\}\) lies in \(d\)-dimensional domain
\(\mathcal{X} \subseteq \mathbf{R}^d\) with sample sizes \(n^{o}\),
cumulative distribution function (CDF) \(F^{o}_{X}\) and probability
density function (PDF) \(f^{o}_{X}\). Let
\(\phi: \mathcal{X} \rightarrow \mathcal{S} \subseteq \mathbf{R}\) be a
score function and define the threshold score \(s \in \mathcal{S}\). The
proportion above the threshold \(s\) in dataset \(o\) is in effect the
contamination rate \(C^{o}(s) = \Pr\left(\phi(x_{i}^o) \ge s\right)\).
The contamination rate is the complementary CDF of the scores,
\(F^{o}(s) = 1 - C^{o}(s)\). As before, \(F^{o}\) and \(f^{o}\) now denote
the CDF and PDF of the scores.

Consider the null hypothesis \(\mathcal{H}_0: F^{te}_{X} = F^{tr}_{X}\)
of equal distribution against the alternative
\(\mathcal{H}_1: F^{te}_{X} \neq F^{tr}_{X}\). In tandem, consider the
null of equal contamination \(\mathcal{H}_0^{s}: C^{te}(s) = C^{tr}(s)\)
against the alternative \(\mathcal{H}_1^{s}: C^{te}(s) \neq C^{tr}(s)\)
at a given threshold score \(s\). To evaluate goodness-of-fit,
\citet{zhang2002powerful} shows that testing
\(\mathcal{H}_0\) is equivalent to testing \(\mathcal{H}_0^{s}\) for
\(\forall s \in \mathcal{S}\). If \(z(s)\) is the relevant test
statistic for equal contamination, a global test statistic \(Z\) for
goodness-of-fit can be constructed from \(z(s)\), its local counterpart.
One such \(Z\) is

\begin{equation}\label{eqn:zhang}
Z = \int z(s) \cdot w(s) \cdot d(s)
\end{equation}

where \(w(s)\) are threshold-dependent weights. For concision, we
sometimes suppress the dependence on the threshold score \(s\) and
denote weights, contamination statistics and rates as \(w\), \(z\) and
\(C^{o}\) respectively. \texttt{D-SOS} differs from the \(Z\) statistic
in Equation (\ref{eqn:zhang}) in three ways: the score function
\(\phi\), the weights \(w\) and the contamination statistic \(z\). We
address each in turn.

\texttt{D-SOS} scores instances from least to most \emph{abnormal}
according to a specified notion of outlyingness. To be concrete, for
density estimation, the negative log density is a natural score for
outlyingness \citep{kandanaarachchi2021leave}. This property
of the score function \(\phi\) can be expressed as

\begin{equation}\label{eqn:phi}
\phi(x_i) \le \phi(x_j) \Rightarrow \Pr(f^{tr}_{X}(x_i) \ge f^{tr}_{X}(x_j)) \ge 1 - \epsilon
\end{equation}

for \(x_i, x_j \in \mathcal{X}\) and \(\epsilon\), a (sufficiently
small) approximation error. Accordingly, instances in high-density
regions of the training set \(X^{tr}\) (nominal points or inliers) score
low; those in low-density regions (outliers) score high. Here, the score
function \(\phi\) can be thought of as a density-preserving projection.
More generally, higher scores, e.g.~wider prediction intervals and
larger residuals, indicate worse outcomes; the higher the score, the
more \emph{unusual} the observation. The structure in \(\phi\) is the
catalyst for adjusting the weights \(w\) and the statistic \(z\).

\texttt{D-SOS} updates the weights \(w\) to be congruent with the score
function \(\phi\). When projecting to the outlier subspace, high scores
imply unusual points, whereas both tails are viewed as extremes in the
Zhang framework. Univariate tests such as the Anderson-Darling and the
Cramér-von Mises tests fit the framework in Equation (\ref{eqn:zhang}):
they make different choices for \(\phi\), \(z\) and \(w\). These
classical tests place more weight at the tails to reflect the severity
of tail exceedances relative to deviations in the center of the
distribution. \texttt{D-SOS} corrects for outliers being projected to
the upper (right) tail of scores via the score function. The
\texttt{D-SOS} weights \(w\) are specified as

\begin{equation}\label{eqn:weights}
w(s) = \left(F^{tr}(s) \right)^2
\end{equation}

The weights in Equation (\ref{eqn:weights}) shift most of the mass from
low to high-threshold regions. As a result, \texttt{D-SOS} is highly
tolerant of negligible shifts associated with low scores, and
conversely, it is attentive to the shifts associated with high scores.
Low (high) thresholds map to high (low) contamination rates and low 
(high) values of the CDF, \(F^{tr}_{X}\). At the lowest threshold, every
point in the training set is an outlier; at the highest threshold, none
is. Within this spectrum, high thresholds better reflect the assumption
that the training set does not contain too many outliers\footnote{We do not
assume that the training set only consists of inliers. It may be contaminated,
but hopefully not too badly compromised i.e.~it is clean enough to be the
reference distribution.}. Thus, we put
more weights on high, rather than low, thresholds. \citet{zhang2002powerful}
posits other functional forms, but the quadratic relationship between weights
and contaminations as in Equation (\ref{eqn:weights}) gives rise to one of
the most powerful variants of the test and so, we follow suit.

\texttt{D-SOS} constructs a test statistic based on the score ranks, not
on the levels. The weighted area under the receiver operating characteristic
curve (WAUC) is a robust statistic that is invariant to changes in levels so
long as the underlying ranking is unaffected. The WAUC, denoted \(T\),
can be written as

\begin{equation}\label{eqn:wauc}
T = \int F^{tr}(s) \cdot f^{te}(s) \cdot w(s) \cdot d(s)
\end{equation}

See Equation (1) in \citet{hand2009measuring} for details. \(T\) in Equation
(\ref{eqn:wauc}) is formally the \texttt{D-SOS} test statistic for adverse
shift. The (W)AUC is also threshold-invariant, meaning that it averages
(discriminative) performance over thresholds of varying importance.
The upside is we do not have to commit to a single threshold.
The downside is, if not careful, the (W)AUC summarizes performances over
irrelevant regions e.g. low-threshold regions. As a generalization of the
AUC, the WAUC puts non-uniform weights on thresholds \citep{li2010weighted}.
\texttt{D-SOS} seizes on this to give more weight to the outlying regions
of the data\footnote{\citet{weng2008new} also advocates for the
WAUC as a performance metric for imbalanced datasets because it can capture
disparate misclassification costs.}. These data-driven (adaptive) weights
conform to the training (reference) distribution, relieving us of the burden
of introducing additional hyperparameters.

The null of no adverse shift \(\mathcal{H}_0^{ds}\) is that most
instances in the test set are not worse than those in the training set.
The alternative \(\mathcal{H}_1^{ds}\) is that the test set contains
more outliers than expected, if the training set is the reference
distribution. \texttt{D-SOS} specifies its null as
\(\mathcal{H}_0^{ds}: T_0 \leq T \mid \mathcal{H}_0\) against the
alternative \(\mathcal{H}_1^{ds}: T_0 > T \mid \mathcal{H}_0\). \(T_0\)
is the observed WAUC and \(T \mid \mathcal{H}_0\), the WAUC under the
null of exchangeable samples. When the test set is cleaner than the
training set, tests of goodness-of-fit and of equal distributions both
reject the null of no difference, whereas \texttt{D-SOS} by design does
not reject its null of no adverse shift. These shifts do not trigger the
alarm as they otherwise would had \texttt{D-SOS} been a two-tailed test.

We make a few remarks to further distinguish \texttt{D-SOS} from other
two-sample statistical tests. Firstly \texttt{D-SOS} does not need to specify
the equivalence margin, the minimum meaningful difference needed to sound the
alarm \citep{wellek2010testing}. This margin is often a prerequisite of
noninferiority tests and similar approaches\footnote{For example, 
\citet{podkopaev2022tracking} sets in advance what drop in accuracy,
say 5\%, constitutes a non-negligible adverse shift.}.
Choosing this margin or minimum effect size is often a non-trivial task,
requiring substantive domain knowledge \citep{betensky2019p}. \texttt{D-SOS}
sidesteps margin elicitation from experts and relies solely on the training
set being an adequate reference distribution. Secondly, tests of equal
distributions ignore that the training set precedes the test set, and not
vice versa. \texttt{D-SOS}, like tests of goodness-of-fit, improve on the
former because they use the training set as the reference distribution.
Lastly, \texttt{D-SOS} deviates from tests of goodness-of-fit. It emphasizes
robustness to fight false alarms, it narrows its scope to detecting adverse
shifts, not any distribution shift, and it accomodates \emph{any} outlier
score\footnote{The score in \citet{zhang2002powerful} is inspired from the
likelihood ratio.}.

\section{Related work}\label{sec:related}

\textbf{Outlier scores}. Density ratios and class probabilities often
serve as scores in comparing distributions \citep{menon2016linking}
These scores, however,
do not directly account for the predictive performance. Prediction
intervals and residuals, on the other hand, do. Intuitively, they both
reflect poor predictive performance in some regions of the feature
space. Confidence-based OOD detection leans on this insight and tracks
\emph{uncertain} predictions e.g. \citet{snoek2019can, berger2021confidence}.
This is in contrast to density-based OOD detection e.g.
\citet{morningstar2021density}. The classical approach, which also accounts
for the predictive performance, is based on residual diagnostics and
underpin misspecification tests. \citet{jankova2020goodness} is a recent
example of this approach with a machine learning twist. Other methods such
as trust scores can also flag unreliable predictions \citep{jiang2018trust}.
Because all these scores represent distinct notions of outlyingness, the
contrasts are often insightful. In
this respect, \texttt{D-SOS} is in some sense a unifying framework.
Bring your own outlier scores and \texttt{D-SOS} would morph into the
equivalent two-sample test for no adverse shift.

\textbf{Dimension reduction}. In practice, the reconstruction errors from
dimension reduction can separate inliers from outliers. Recently,
\citet{rabanser2019failing} uses dimension
reduction, as a preprocessing step, to detect distributional shifts.
Some methods gainfully rely on the supervised predictions for their low-rank
representation \citep{cieslak2009framework, lipton2018detecting}. This last
approach, to add a cautionary note, entangles the supervised model,
subject to its own sources of errors such as misspecification and overfitting,
with dataset shift \citep{wen2014robust}. But it also points to the
effectiveness of projecting to a lower and more informative subspace to
circumvent the curse of dimensionality. Indeed, the classifier
two-sample test uses \emph{univariate} scores to detect changes in
\emph{multivariate} distributions \citep{cai2020two}.
Inspired by this approach, \texttt{D-SOS} uses outlier scores as a
device for dimension reduction.

\textbf{Statistical tests}. The area under the receiver operating
characteristic curve (AUC) has a long tradition of being used as a robust test
statistic in two-sample comparison. \citet{demidenko2016p} proposes an AUC-like
test statistic as an alternative to the classical tests at scale because the
latter ``do not make sense with big data: everything becomes
statistically significant'' while the former attenuates the strong bias
toward large sample size. \texttt{D-SOS} is firmly rooted in
the tradition of two-sample comparison. Inference is at the sample
level. At the instance level, we may want to check if a given test point
is an outlier to the training set. \citet{bates2021testing} for example tests
for individual outliers via conformal \(p-\)values. Turning pointwise outlier
score into an interpretable (standardized) measure such as a probability is a
common and useful desideratum \citep{kriegel2009loop, kriegel2011interpreting}.
\citet{podkopaev2022tracking}, on the other hand, casts the problem of
detecting harmful shifts in a sequential setting and controls for the
false alarm rate due to continuous monitoring. In sequential settings,
statistical process control e.g. \citep{qiu2020big} is often used for a
similar purpose.

\section{Implementation}\label{sec:implementation}

In this section, we turn our attention to deriving valid \(p-\)values.
Without loss of generality, let \(\mathcal{D} = [X^{tr}, X^{te}]\) be
the pooled training and test set. The score function
\(\phi = \Phi(\mathcal{D}, \lambda)\) is estimated from data
\(\mathcal{D}\) and hyperparameters \(\lambda \in \Lambda\). This
calibration procedure
\(\Phi: \mathcal{X} \times \Lambda \rightarrow \phi\) returns the
requisite score function
\(\phi: \mathcal{X} \rightarrow \mathcal{S} \subseteq \mathbb{R}\). The
asymptotic null distribution of the WAUC, however, is invalid when the
same data is used both for calibration and scoring. We circumvent this
issue with permutations, sample splitting, and/or out-of-bag
predictions.

Permutations use the empirical, rather than the asymptotic, null
distribution. The maximum number of permutations is set to \(R = 1000\).
We refer to this
variant as \texttt{DSOS-PT}. Unless stated otherwise, this is the default
used in Section \ref{sec:experiments}. For speed, \texttt{DSOS-PT} is
implemented as a sequential Monte Carlo test, which terminates early
when the resampling risk is unlikely to affect the final result materially
\citep{gandy2009sequential}. Even so, permutations can be computationally
prohibitive.

A faster alternative, based on sample splitting, relies on the asymptotic
null distribution. It incurs a cost in calibration accuracy in the first step
because it holds out half the data for scoring in the second step.
This tradeoff is common in
two-sample tests, which requires calibration. We randomly
split each dataset in half: \(X^{tr} = X_1^{tr} \cup X_2^{tr}\) and
\(X^{te} = X_1^{te} \cup X_2^{te}\). The first halves are used for
calibration, and the second for scoring. We describe this split-sample
procedure in Algorithm \ref{algo:dsos-2}, and refer to it as
\texttt{DSOS-SS} in Section \ref{sec:experiments}. Given the weights in
Equation \ref{eqn:weights}, the WAUC under the null
\(T\mid \mathcal{H}_0\) is asymptotically normally distributed as

\begin{equation}\label{eqn:null}
	T \mid \mathcal{H}_0 \sim \mathcal{N}(\frac{1}{12}, \sigma(n^{tr}, n^{te}))
\end{equation}

with mean \(= \frac{1}{12}\) and standard deviation
\(\sigma(n^{tr}, n^{te})\), which depends on the sample sizes.
Equation \ref{eqn:null} follows trivially from results in
\citet{li2010weighted} with a little bit of
tedious algebra -- see Equation (9) therein for the variance.

A third option, an obvious extension to sample splitting, is to use
cross-validation instead. In \(k-\)fold cross-validation, the number of
folds \(k\) mediates between calibration accuracy and inferential
robustness. At the expense of refitting \(k\) times, this approach uses
most of the data for calibration and can leverage the asymptotic null
distribution, provided that the scores are out-of-sample predictions.
This strategy is also called cross-fitting in semiparametric inference.
It combines the best of both worlds, namely calibration accuracy and
inferential speed. We refer to this variant as
\texttt{DSOS-CV} in Section \ref{sec:experiments}. We show in
simulations that this approach either matches or exceeds the performance
of \texttt{DSOS-PT} and \texttt{DSOS-SS}.

\begin{algorithm}
	\SetAlgoLined
	\KwData{Calibration set $\mathcal{D}_1 = [X_1^{tr}, X_1^{te}]$,
		Inference set $\mathcal{D}_2 = [X_2^{tr}, X_2^{te}]$, calibration procedure
		$\Phi$ and hyperparameters $\lambda$}
	\SetKwFunction{WeightedAUC}{WeightedAUC}
	\BlankLine
	
	\textbf{Label:} Assign labels $y = \mathbb{I}(x \in X_2^{te})$,
	$\forall x \in \mathcal{D}_2$\;
	\textbf{Calibrate:} Fit the score function
	$\phi = \Phi(\mathcal{D}_1, \lambda)$\;
	\textbf{Score:} Score observations $s = \phi(x)$, $\forall x \in \mathcal{D}_2$\;
	\textbf{Test:} Compute the observed WAUC
	$T_0 = \WeightedAUC(s, y)$ using Equation (\ref{eqn:wauc})\;
	\textbf{Asymptotic null:} Under the null, $T$ is asymptotically
	normally distributed as in Equation (\ref{eqn:null})\;
	\BlankLine
	\KwOut{$p-value = 1 - \Pr(T \le T_0)$}
	\caption{Split-sample test (DSOS-SS)}
	\label{algo:dsos-2}
\end{algorithm}

\section{Experiments}\label{sec:experiments}

We make the following pragmatic choices for ease of use; typically, the
selected score functions perform well out-of-the-box with little to no
costly hyperparameter tuning. For density-based OOD detection, we use
isolation forest. To investigate other notions of outlyingness, we use random
forests. As in \citet{hediger2022use}, random
forests allow us to use the out-of-sample variant of \texttt{D-SOS}
(\texttt{DSOS-CV}) for free, so to speak. Out-of-bag predictions
are viable surrogates for out-of-sample scores. This out-of-bag variant is,
when feasible, often convenient as it improves on sample splitting, which
sacrifices calibration accuracy for inferential robustness. Details
about the experimental setup are in the supplementary material.

\begin{table}
    \centering
    \caption{Number of wins for test comparisons based on simulations. 
	\textsuperscript{1}Number of ties, not shown, when s-value difference is
	within ROPE.}\label{tab:simulations}
    \begin{tabular}{llrr}
      \toprule 
      \multicolumn{2}{c}{Contender} & \multicolumn{2}{c}{Outcome\textsuperscript{1}} \\
      \cmidrule(lr){1-2} \cmidrule(lr){3-4}
	(1) & (2) & Win (1) & Win (2) \\
      \midrule 
	ctst & energy & 12 & 43 \\
	ctst & DSOS-SS & 0 & 34 \\
	ctst & DSOS-PT & 0 & 60 \\
	ctst & DSOS-CV & 0 & 64 \\
	DSOS-CV & energy & 37 & 26 \\
	DSOS-SS & DSOS-CV & 0 & 36 \\
	DSOS-SS & DSOS-PT & 0 & 32 \\
	DSOS-CV & DSOS-PT & 0 & 0 \\
      \bottomrule 
    \end{tabular}
\end{table}

\subsection{Simulated shifts}\label{exp:simulation}

We compare \texttt{D-SOS} to two modern tests of equal distribution in
unsupervised settings. These benchmarks illustrate the simplicity and
performance of our approach where we have an abundance of choice. The field
has been churning out new and powerful two-sample tests of equal distribution.
These tests are a natural baseline for \texttt{D-SOS} because they are commonly
used to detect distribution shift. We look at both the \texttt{energy} and
the classifier test. We show that overall, \texttt{D-SOS} performs
favorably. We lose little by using a test for no adverse shift instead
of a test of equal distribution. \texttt{D-SOS} seldom loses 
and often matches or exceeds the performance of these tests.

The first test is the \texttt{energy} test, a type of kernel-based (distance-based)
test \citep{szekely2004testing}. The second, arguably the main inspiration
for \texttt{D-SOS}, is a classifier two-sample test, \texttt{ctst}
\citep{kim2021classification}. \texttt{ctst} tests whether a classifier
can reliably identify training from test instances. If so, it is taken
as evidence against the null. Both \texttt{D-SOS} and \texttt{ctst} leverage
the same underlying classifier and classification scores. As this score
increases, so does
the likelihood that an observation belongs to the test set, as opposed
to the training set. Following \citet{ciemenccon2009auc}, this \texttt{ctst}
variant uses sample splitting, as does \texttt{DSOS-SS}, but the AUC
rather than the WAUC as a test statistic. The weighting scheme is what
differentiates the two. 

We simulate distribution shifts from a two-component multivariate
Gaussian mixture model and specify distinct shift type (setting): no
shift (the base case), label shift, corrupted sample, mean shift, noise shift,
and dependency shift. This is described in detail in the supplementary material.
When applicable, the shift intensity toggles between low, medium, and high for
each shift type. We also vary sample size and dimension. For each combination of
shift type, shift intensity, sample size, and dimension, we repeat the
experiment 500 times.

To compare the \texttt{D-SOS}, \texttt{ctst} and \texttt{energy} tests, we
employ the Bayesian signed rank test \citep{benavoli2014bayesian}. To do
so, we specify a region of practical equivalence (ROPE) on the \(s\)-value
scale and a suitable prior. We deem two statistical tests practically
equivalent if the absolute difference in \(s\)-value is \(\leq 1\). This
difference corresponds to one more coin flip turning up as heads (or
tails), keeping the streak alive. Anything greater arouses suspicion
that one test is more powerful, and that the difference is not
negligible. Specifying a ROPE on the \(s\)-value scale is less
cumbersome than the \(p\)-value scale. \(p\)-values, let alone
\(p\)-value differences, are notoriously difficult to interpret
\citep{wasserstein2016asa}. The Bayesian signed
rank test yields the posterior probability that one method is better,
practically equivalent or worse than the other. The prior for these
experiments adds one \emph{pseudo} experiment, distinct from the 500
\emph{real} simulations, where the difference is 0, i.e.~no practical
difference exists out of the gate.

Table \ref{tab:simulations} summarizes the
findings across all settings: dimension, sample size, shift type and
shift intensity. The full tables for comparison are provided as
supplementary material. For simplicity, we say that one method wins if the
posterior probability that it is better is \(\geq 0.5\); similarly, they
draw if the posterior probability that they are practically equivalent
is \(\geq 0.5\). This way of comparing methods resembles the Pitman
closeness criterion. We make several observations:

\begin{enumerate}
	\def\labelenumi{\arabic{enumi}.}
	\item
	\texttt{DSOS-PT} and \texttt{DSOS-CV} dominate \texttt{DSOS-SS}.
	\texttt{DSOS-SS} pays a hefty price in inferential accuracy due to
	sample splitting. Sample splitting, as evidenced in \texttt{DSOS-SS}
	and \texttt{ctst}, clearly trails behind the competition. Even so,
	\texttt{DSOS-SS} outperforms this \texttt{ctst} variant: the weights
	in \texttt{DSOS-SS} clearly pay dividends. The WAUC improves upon the
	plain AUC and offers a compelling alternative to the Mann-Whitney-Wilcoxon
	(AUC-based) tests when robustness is a chief concern.
	\item
	\texttt{DSOS-CV} (or \texttt{DSOS-PT}) lag behind \texttt{energy} in
	settings with (1) label shift and (2) corrupted samples. There are 26
	such cases. This is because \texttt{DSOS} is robust to
	outliers and more forgiving than nonrobust tests when the bulk of
	the distributions is largely unaffected. In all other configurations,
	\texttt{DSOS-CV} either wins or draws. In particular, there are 37 cases
	where \texttt{DSOS-CV} beats \texttt{energy}. This is because, like tests
	of goodness-of-fit, \texttt{D-SOS} benefits from conditioning on the
	the reference distribution while being robust to outliers.
	\item
	When there is indeed a shift, power (\(s\)-value) increases when sample
	size and shift intensity increases and decreases when dimension
	increases, all else equal. This trend is consistent and is as expected.
	However, the relative efficiency of these tests in small samples and/or at
	low shift intensity can at times be striking. The 26 cases where
	\texttt{D-SOS} loses to the \texttt{energy}	test can be characterized as
	such. In large samples and/or high shift intensity, asymptotics kick in
	and they all converge.
	\item
	\texttt{DSOS-PT} and \texttt{DSOS-CV} are practically equivalent
	across all settings. Notwithstanding the gains in inferential speed
	that one may hold over the other, the inferential accuracy (power) is
	-- should be -- the same. This is a good sanity check that the
	implementation is correct. Similarly, for the simulations with no 
	distribution shift, all tests are practically equivalent.
\end{enumerate}

\begin{figure*}
	\centering
	\includegraphics[width=\linewidth]{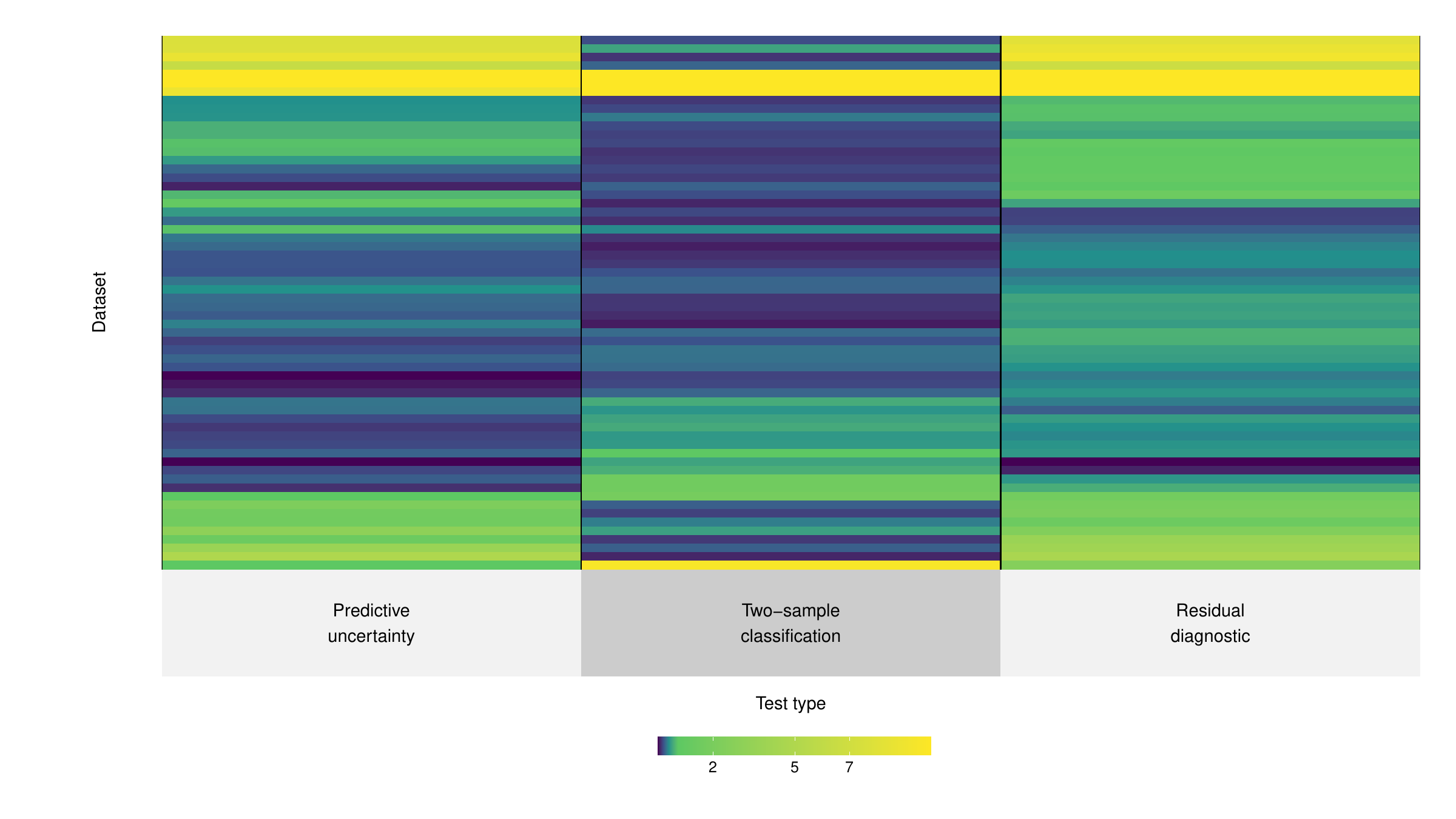}
	\caption{
		Heatmap of fixed effects for 62 datasets from the OpenML-CC18 benchmark,
		ordered via hierarchical clustering. Fixed effects are exponentiated and
		reported on the $s$-value scale for 3 types of \texttt{D-SOS} test. The
		brighter the color, the higher the $s$-value, the more prone the dataset
		to adverse shift caused by sampling variation.}
	\label{fig:heatmap-cc18}
\end{figure*}

\subsection{Partition-induced shifts}\label{exp:cc18}

In supervised settings, since the core task \emph{is} prediction, the 
\texttt{D-SOS} tests based on residual diagnostics or predictive uncertainty
are more informative than, and can be at odds with, classifier (density-based)
tests. \texttt{D-SOS} creates rigorous statistical tests from these scores
so that they can be held to the same standard of evidence as null
hypothesis statistical test. This experiment scales beyond the
motivational -- admittedly, somewhat contrived -- example in Section
\ref{sec:iris} and shows that those findings extend to and are stable across
datasets.

To investigate how different notions of outlyingness are correlated, we analyze
via cross-validation 62 classification tasks from the OpenML-CC18 benchmark
\citep{OpenMLR2017}\footnote{We exclude
datasets with more than 500 features because of long runtimes.}. As in
Section \ref{sec:iris}, we look at tests of no adverse shift based on
two-sample classification, residual diagnostics, and predictive
uncertainty (no density-based OOD detection in this round). Cross-validation
mimics random sampling variation, and is helpful in assessing stability 
\citep{moreno2012study, lim2016estimation}; it generates partition-induced
shifts, rather than worst-case (adversarial) perturbations
\citep{subbaswamy2021evaluating}.

For each dataset, we estimate a fixed (mean) effect reported on the \(s-\)value
scale, which measures how sensitive a dataset is to these partition-induced
shifts. The higher the value, the more susceptible a dataset is to adverse shifts
caused by sampling variation. These fixed effects can be interpreted in the
usual way as the strength of evidence against the null of no adverse shift. We
refer to the supplementary material for further details about this experiment.
At a glance, Figure \ref{fig:heatmap-cc18}, created with the \texttt{superheat}
package \citep{barter2018superheat}, tells the story quite succintly.

Across these 62 datasets, we see that tests of no adverse shift based on
two-sample classification, residual diagnostics, and predictive uncertainty
(confidence-based OOD detection) are indeed highly correlated. Residual
diagnostics and predictive uncertainty, both of which mirror the performance
of the underlying supervised algorithm, are the most correlated (Pearson
correlation coefficient is 0.82). This suggests that they can be used
interchangeably, and by virtue of focusing on predictive performance, are
suitable metrics for model monitoring. Two-sample classification is not as
strongly associated with residual diagnostics and predictive uncertainty
(Pearson correlation coefficient is 0.5 and 0.47 respectively). This arcs
back to the point made in Section \ref{sec:iris}; namely, tests of equal
distribution alone are ill-equiped, by definition, to detect whether a shift
is benign or harmful for predictive tasks.

Notice the datasets clustered together at the very top of Figure
\ref{fig:heatmap-cc18} with low \(s-\)values for two-sample
classification but high \(s-\)values for residual diagnostics and
predictive uncertainty. For these datasets, the distribution seems fine
but prediction suffers. To a somewhat lesser extent, also notice the reverse
effect at the bottom of Figure \ref{fig:heatmap-cc18}. These are the
false alarms we previously highlighted where we have relatively higher
\(s-\)values for two-sample classification than for the other two
notions of outlyingness. For these datasets, prediction seems fine but the
distribution is not. The link between distribution shift and predictive
performance can be more tenuous than what we typically assume. When faced with
dataset shift, by rote we often reach for tests of equal distribution. It
may be more productive to deliberate over the appropriate notions of
outlyingness to detect adverse shift.

\section{Conclusion}\label{sec:conclusion}

\texttt{D-SOS} is a wrapper to turn outlier scores into a statistical
test. This framework, derived from tests of goodness-of-fit, detects
non-negligible adverse shifts. It can confront the data with more
pertinent hypotheses than tests of equal distribution. This works well
when mapping to the relevant outlier subspace discriminates between safe
and unsafe regions of the data distribution. Our method accommodates
different notions of outlyingness. It lets users define, via a score
function, what is meant by \emph{worse} off. Besides the outlier scores
explored in this paper, we stress that other sensible choices for the
score functions \(\phi\) and the weights \(w\) abound. These can be
adjusted to account for domain knowledge. 
Looking ahead, future research could investigate other
weighting schemes. The functional form of the postulated weights could
be worth tuning. Moreover, \emph{composite} score function, which would
combine several notions of outlyingness together, would enrich the types
of hypotheses we can test. We imagine situations where we want to
combine the strengths of confidence-based and density-based OOD
detection into a single aggregate score for example.

\begin{acknowledgements} 
	The author would like to thank the reviewers for suggesting numerous
	improvements. Sukanya Srichandra was an excellent sounding board. This
	work was supported by the Royal Bank of Canada (RBC). The views expressed
	here are those of the author, not of RBC.
\end{acknowledgements}

\bibliography{kamulete_5.bib}

\end{document}